\documentclass[10pt,twocolumn,letterpaper]{article}

\usepackage[algorithms]{wacv}

\usepackage{amsmath,amssymb,amsfonts}
\usepackage{algorithm}
\usepackage{algorithmic}
\usepackage{graphicx}
\usepackage{xcolor}
\usepackage{booktabs}
\usepackage{multirow}
\usepackage{array}
\usepackage{microtype}
\graphicspath{{figures/}}
\usepackage[colorlinks=true,allcolors=blue]{hyperref}
\usepackage[capitalize]{cleveref}
\crefname{section}{Sec.}{Secs.}
\Crefname{section}{Section}{Sections}
\Crefname{table}{Table}{Tables}
\crefname{table}{Tab.}{Tabs.}

\def\wacvPaperID{}
\def\confName{WACV}
\def\confYear{2027}

\newcommand{\Lcls}{\mathcal{L}_{\text{cls}}}

\newcommand{\Lfpr}{\mathcal{L}_{\text{FPR}}}
\newcommand{\bx}{\mathbf{x}}
\newcommand{\bw}{\mathbf{w}}
\newcommand{\E}{\mathbb{E}}

\providecommand{\etal}{\textit{et al.}}
\newcommand{\mypar}[1]{\vspace{2pt plus 1pt minus 1pt}\noindent\textbf{#1.}~}
\newcommand{\good}[1]{{\color{teal}#1}}
\newcommand{\bad}[1]{{\color{red!70!black}#1}}
\newcommand{\cmark}{{\color{teal}\checkmark}}
\newcommand{\xmark}{{\color{red!70!black}$\times$}}

\begin{document}

\title{FYC: FPR-Disparity Weight Perturbation and Youden Calibration\\
for Fair Deepfake Detection}

\author{Uzoamaka Ezeakunne$^{1}$\quad Taiwo Olabamipe$^{2}$\quad Deevesh Jain$^{1}$\quad Xiuwen Liu$^{1}$\\
$^{1}$Department of Computer Science, Florida State University\\
$^{2}$Department of Computer Science, University of Central Florida\\
{\tt\small uezeakunne@fsu.edu, ta117847@ucf.edu, dj23f@fsu.edu, xliu@fsu.edu}
}

\maketitle

\begin{abstract}
Deepfake detectors routinely achieve high aggregate accuracy while
exhibiting substantial demographic disparities in their errors:
false-positive rates differ by double digits across gender and skin
tone. Most existing remedies require either full retraining with a
fairness-aware objective or demographic-annotated training data, both
impractical when only a pretrained checkpoint is available, while
label-free post-hoc calibrators leave the detector's weights unchanged.
We introduce \textbf{FYC} (FPR-disparity Weight Perturbation and Youden
Calibration), a two-stage post-hoc method that perturbs a detector's
weights under an FPR-disparity criterion, then calibrates a separate
decision threshold for each demographic group---requiring no
architectural changes and no full retraining. Evaluated against every
detector benchmarked on AI-Face v2 across six intersectional gender
$\times$ skin-tone groups, FYC outperforms the field on 18 of 21
reported fairness and utility metrics, including purpose-built fairness
architectures, at a fraction of the training cost.
We further identify why the choice of fairness criterion is critical:
the original Demographic Parity objective used in prior weight-perturbation
work is architecture-dependent and unpredictable, improving two of four
evaluated architectures, leaving a third largely unchanged, and
substantially degrading the fourth (PG-FDD), with no way to anticipate
the outcome for a given checkpoint in advance. Conditioning the same
objective on real images only---our FPR-disparity criterion---eliminates
this unpredictability, improving fairness and accuracy jointly across
all four architectures with no exceptions.
\end{abstract}

\section{Introduction}
\label{sec:intro}

Modern deepfake detectors achieve impressive aggregate accuracy yet
conceal striking demographic disparities.
Trinh and Liu~\cite{trinh2021examining} report that a detector's
false-positive rate is nearly twice as high for female vs.\ male subjects
(14.0\% vs.\ 7.7\%), and detection accuracy may vary by over 40\% across
intersectional subgroups~\cite{xu2024analyzing}.
From a deployment perspective, biased detectors disproportionately
flag authentic content from minority groups, interfere with identity
verification for certain demographics, and create exploitable
blind spots for adversarial manipulation~\cite{trinh2021examining,xu2024analyzing}.

Existing remediation approaches fall into four categories
(Table~\ref{tab:methods}).
\emph{Pre-processing} methods balance training data but require
significant annotation effort and may discard valuable
samples~\cite{nadimpalli2022gbdf}.
\emph{In-processing} methods such as DAG-FDD~\cite{ju2023improving} and
PG-FDD~\cite{lin2024preserving}
modify training objectives or architectures, but require full
retraining and, in most cases, explicit demographic
annotations---requirements that conflict with production release
cycles and the use of commercial pretrained models.
\emph{Post-hoc weight adaptation} methods such as weight
perturbation~\cite{jiang2023mwp} fine-tune an already-trained model's
weights without full retraining or architecture changes, starting from
a frozen pretrained checkpoint.
\emph{Post-processing} methods adjust outputs or decision boundaries after
training without touching weights at all; per-group threshold
calibration~\cite{hardt2016equality} is the simplest such approach but
does not modify the underlying discriminative geometry of the detector.

\mypar{This work}
Weight perturbation---an algorithm that ascends toward worst-case
group-disparity configurations before descending with a
fairness-regularized loss---retrofits fairness into pretrained
deepfake detectors without architectural changes.
The choice of disparity criterion determines how \emph{reliable} the
outcome is.
Its original Demographic Parity (DP) objective---parity satisfied by
equalizing positive-prediction rates rather than correct predictions,
without conditioning on the true label---is architecture-dependent and
unpredictable: it improves intersectional fairness for two of our four
evaluated architectures, is roughly flat for a third, and substantially
worsens it only for the fourth (PG-FDD, on the skin-tone axis
specifically), with no way to know in advance which case a given
checkpoint will land in (\S\ref{sec:dp_diagnostic}).
Conditioning the same objective on real images only (FPR-disparity)
removes this unpredictability: it improves both intersectional
fairness and AUC directly for all four architectures, with no
exception; a zero-cost per-group Youden-J calibration stage then adds
further gains on top.
We call this two-stage recipe \textbf{FYC} (FPR-disparity Weight
Perturbation and Youden Calibration).
FYC is practical---a single GPU, a few hours, no architectural
changes---and outperforms \emph{all 12} detectors benchmarked on
AI-Face v2~\cite{aiface}, including purpose-built fairness architectures
that require full retraining, on 18 of 21 reported fairness and utility
rows (\S\ref{sec:sota}).

\mypar{Contributions}
(1) \textbf{New best result on AI-Face v2}: FYC, applied to Xception,
outperforms all 12 AI-Face-benchmarked detectors---including the
purpose-built fairness architecture PG-FDD and the CLIP-based
UnivFD---on 18 of 21 fairness and utility rows (\S\ref{sec:sota},
Table~\ref{tab:sota}), requiring only ${\approx}10$--$15$\,h of post-hoc
fine-tuning (39\,min per epoch) versus
full retraining for every baseline.
Within our own four-architecture suite, FYC reduces intersectional
FMEO by 38.0--70.1\% and global FPR by 78.2--91.5\%, with AUC
improved for all four.
(2) \textbf{Diagnostic finding}: DP-regularized weight perturbation's
raw-stage effect on intersectional fairness is architecture-dependent
rather than uniformly destructive---it improves two of our four
evaluated architectures and is roughly flat for a third, but
substantially worsens the fourth (PG-FDD, with the degradation
concentrated on the skin-tone axis)---making its raw stage an unreliable default a
practitioner cannot vet without running the diagnostic itself, and
motivating the uniformly-reliable FPR-disparity criterion used in (1).
(3) \textbf{Loss-criterion analysis}: conditioning the disparity objective
on real images ($y{=}0$) removes this unpredictability; FPR-disparity
weight perturbation improves both AUC and intersectional FMEO at
the raw (pre-calibration) stage for \emph{all four} architectures, with
no exception (\S\ref{sec:results}).
(4) \textbf{Comprehensive evaluation} on AI-Face v2 (329,309 test images,
6 intersectional groups) using the dataset author's evaluation pipeline.

\section{Related Work}
\label{sec:related}

\subsection{Fairness in Deepfake Detection}

Trinh and Liu~\cite{trinh2021examining} first systematically documented
demographic disparities across popular detectors, reporting large
error-rate gaps across racial groups and noting that training sets are
overwhelmingly composed of Caucasian subjects.
Xu \etal~\cite{xu2024analyzing} confirmed substantial dataset skew and
detector sensitivity to attributes unrelated to manipulation across a
massively annotated benchmark, and Liu \etal~\cite{liu2025racialbias}
contribute a dedicated race-focused forgery-detection benchmark and
evaluation protocol.
The scale of this problem motivated a dedicated NeurIPS 2025 competition
(2{,}100+ submissions, 162 teams)~\cite{hu2026aifacecompetition}.
We focus throughout on \emph{group} fairness across the six
intersectional gender$\times$skin-tone groups; individual-fairness
formulations for deepfake detection, where similar inputs should
receive similar outputs, are explored concurrently by
Hou \etal~\cite{hou2025individualfairness} and are outside our scope.

\subsection{Mitigation Strategies}

\mypar{Pre-processing}
Methods that balance or augment training data can reduce demographic
skew but require manual annotation effort and may discard informative
samples~\cite{nadimpalli2022gbdf}.
Synthetic-face augmentation offers demographic control but exhibits
a persistent domain gap and does not substitute for real-identity
diversity.

\mypar{In-processing}
\textit{DAG-FDD / DAW-FDD}~\cite{ju2023improving} introduce CVaR-based
losses trained from the labeled data directly: the demographic-agnostic
variant (DAG) minimizes worst-case latent group risk; the
demographic-aware variant (DAW) nests CVaR across specified groups.
Because the fairness objective is baked into the training loss itself,
neither can be applied post-hoc to an already-deployed, frozen
checkpoint without retraining from that point.
\textit{PG-FDD}~\cite{lin2024preserving} disentangles demographic,
domain-agnostic, and domain-specific forgery features and trains with
a distribution-aware fairness loss plus Sharpness-Aware Minimization,
targeting cross-domain fairness generalization specifically---an
important complementary goal, but one that still requires jointly
training the disentangled feature representation from scratch, not
adapting an existing pretrained detector.
\textit{Weight Perturbation}~\cite{jiang2023mwp} generalizes SAM~\cite{foret2021sam}
by replacing the sharpness objective with a DP disparity objective:
it ascends toward the worst-case fairness-violating weight configuration,
then descends with a DP-regularized loss; we build directly on this
mechanism in \S\ref{sec:mwp} but replace its disparity criterion.
\textit{Decoupling Bias}~\cite{ding2025decoupling} instead
architecturally channel-decouples demographic-sensitive features and
aligns their distributions across groups, while \textit{DAID}~\cite{cheng2025daid}
pursues a causal fairness--generalization framework via demographic-aware
data rebalancing; a related SAM-based approach~\cite{ezeakunne2025icaart}
ascends on generic sharpness rather than a disparity gradient and adds
synthetic data balancing with no calibration stage (both evaluated on
non-AI-Face benchmarks).
FairReL~\cite{lu2026fairrel} and FairForensics~\cite{zhang2026fairforensics}
learn fairness-aware representations, the latter with vision-language
supervision; both are trained end to end rather than applied to an
existing checkpoint.
All of these methods require training a model (or a substantial
component of one) from scratch on labeled data, not adapting an
existing pretrained checkpoint.

\mypar{Post-hoc weight adaptation}
A fourth category, distinct from both in-processing (train from scratch)
and post-processing (touch outputs only), fine-tunes an already-trained
model's weights without full retraining or architecture changes.
\textit{BMFT}~\cite{xue2024bmft} identifies and masks the specific
weights most responsible for biased predictions, then fine-tunes in two
stages on a frozen medical-imaging classifier---a sibling to weight
perturbation in this category that intervenes by \emph{masking} rather
than \emph{perturbing} weights.
FYC's Stage 1 (\S\ref{sec:mwp}) belongs to this same category: it
starts from a frozen pretrained checkpoint, requires no architecture
change, and needs only a few hours of fine-tuning rather than a full
training run.

\mypar{Post-processing}
\textit{Youden-J calibration}~\cite{hardt2016equality} finds per-group
optimal decision thresholds that maximize $\text{TPR}_g - \text{FPR}_g$
on a held-out calibration split; it requires group labels at calibration
time but is fully retraining-free and touches outputs only, never
weights.
More recent general-purpose frameworks generalize this idea:
LinearPost~\cite{xian2024linearpost} unifies statistical-parity,
equal-opportunity, and equalized-odds post-processing into a single
linear-program formulation, and FRAPPÉ~\cite{tifrea2024frappe} converts
arbitrary in-processing fairness regularizers into post-processing
procedures.
\textit{Fair-FLIP}~\cite{szandala2025fairflip} is a direct
deepfake-detection-specific competitor: it reweights a trained
detector's final-layer inputs by subgroup variability---closer to our
Stage 2 in spirit but operating on activations rather than thresholds.
Concurrent work, \textit{FFT}~\cite{brown2026fft}, learns a
demographic-label-free residual calibrator from frozen face embeddings
but, by its own numbers, reduces equalized-odds violation only
marginally in-domain and worsens it under distribution shift, consistent
with our finding that weight modification, not calibration alone,
drives most of the fairness gain (\S\ref{sec:cal_only}).

\subsection{Positioning}

Table~\ref{tab:methods} situates our approach across all four
categories.
FYC combines the third and fourth: Stage 1 is post-hoc weight
adaptation (reshaping the discriminative geometry of an already-trained
model), Stage 2 is classical post-processing (correcting the per-group
threshold) on top of it.
No prior method in either category combines the two: post-processing
methods (Youden-J, FFT, Fair-FLIP, LinearPost, FRAPPÉ) never touch
weights, and post-hoc weight-adaptation methods (weight perturbation,
BMFT) are evaluated without a calibration stage on top.
Unlike in-processing methods, FYC requires only a pretrained checkpoint
plus a labeled calibration split; no full retraining and no
architecture change is needed at any stage.

\begin{table}[t]
\caption{Landscape of fair deepfake detection methods.
Label-free: no demographic labels required at any stage.
Weight-free: no model weight updates.}
\label{tab:methods}
\resizebox{\linewidth}{!}{%
\begin{tabular}{llcc}
\toprule
\textbf{Method} & \textbf{Category} & \textbf{Label-free} & \textbf{Weight-free} \\
\midrule
DAG-FDD~\cite{ju2023improving}       & In-processing  & \cmark & \xmark \\
DAW-FDD~\cite{ju2023improving}       & In-processing  & \xmark & \xmark \\
PG-FDD~\cite{lin2024preserving}      & In-processing  & \xmark & \xmark \\
DAID~\cite{cheng2025daid}            & In-processing  & \xmark & \xmark \\
\midrule
Weight Perturbation~\cite{jiang2023mwp} & Post-hoc weight adapt.  & \xmark & \xmark \\
BMFT~\cite{xue2024bmft}              & Post-hoc weight adapt.  & \xmark & \xmark \\
\midrule
Youden-J cal.~\cite{hardt2016equality} & Post-processing & \xmark & \cmark \\
Fair-FLIP~\cite{szandala2025fairflip} & Post-processing & \xmark & \cmark \\
FFT~\cite{brown2026fft}              & Post-processing & \cmark & \cmark \\
\midrule
\textbf{FYC (ours)}                  & Weight adapt.\,+\,post-proc. & \xmark & \xmark \\
\bottomrule
\end{tabular}}
\end{table}

\section{Method}
\label{sec:method}

\begin{figure*}[t]
\centering
\includegraphics[width=0.92\linewidth]{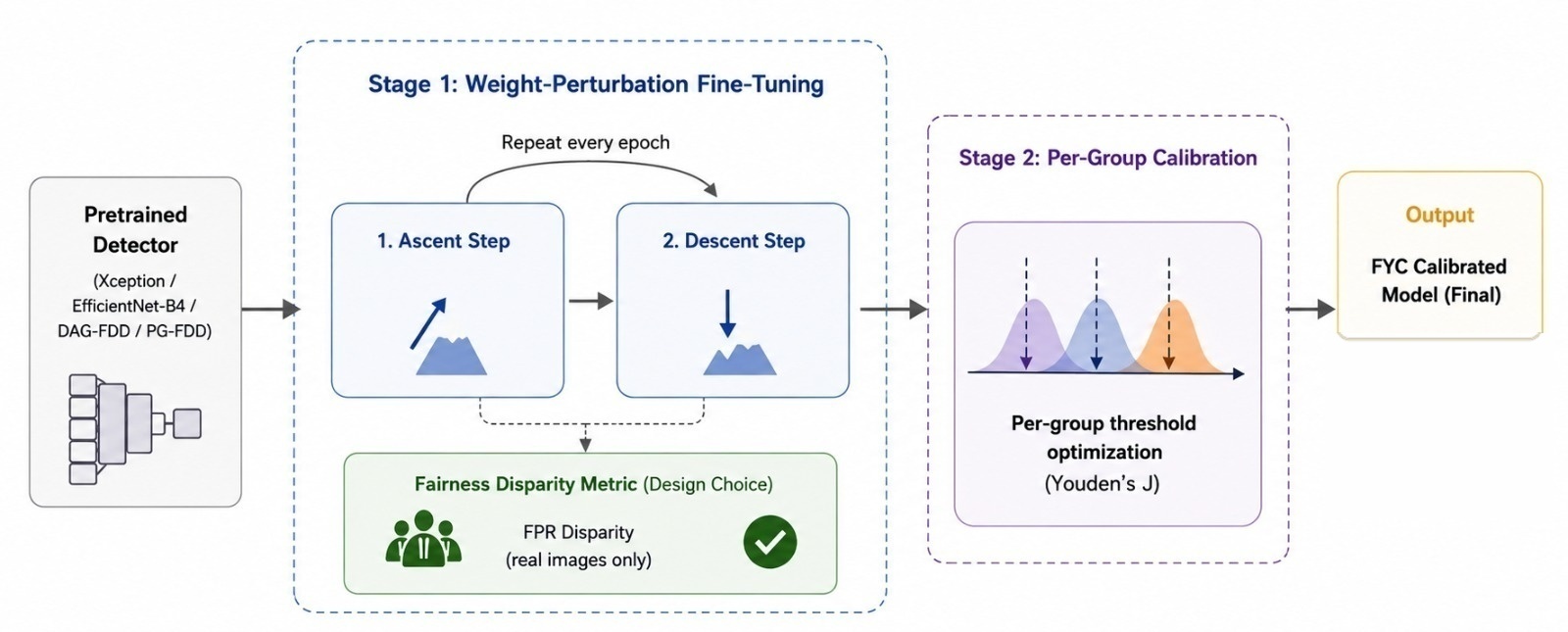}
\caption{Method overview. Stage 1 fine-tunes a pretrained detector with
a weight-perturbation ascent/descent loop, using FPR-disparity (real
images only) as the disparity criterion driving the ascent---the
criterion we adopt after diagnosing it as reliable across all four
architectures we test, unlike the Demographic Parity alternative
(\S\ref{sec:dp_diagnostic}). Stage 2 applies zero-cost per-group
Youden-J calibration (\S\ref{sec:cal}) on top of the resulting
checkpoint, producing the final calibrated model.}
\label{fig:pipeline}
\end{figure*}

\subsection{Problem Formulation}

Let $\mathcal{D} = \{(\bx_i, y_i, g_i)\}_{i=1}^n$ be a dataset of face
images with labels $y_i \in \{0,1\}$ (real/fake) and demographic group
labels $g_i \in \mathcal{G}$.
A detector $f_\theta : \mathcal{X} \to [0,1]$ outputs fake probabilities.
We quantify fairness via:
\begin{align}
\Delta\text{TPR} &= \max_g \text{TPR}_g - \min_g \text{TPR}_g, \\
\text{FMEO} &= \max(\Delta\text{TPR}, \Delta\text{FPR}), \\
\text{EO} &= \Delta\text{TPR} + \Delta\text{FPR},
\end{align}
where $\Delta\text{FPR}$ is the analogous false-positive rate gap.
The AI-Face benchmark~\cite{aiface} implements FMEO and FEO ($=\text{EO}$)
via its \texttt{acc\_fairness()} function (\S\ref{sec:setup}).
We additionally report AUC, global Acc, and global FPR as utility metrics.

\subsection{Stage 1: Weight-Perturbation Fine-Tuning}
\label{sec:mwp}

Figure~\ref{fig:pipeline} overviews both stages of the recipe.
Jiang \etal~\cite{jiang2023mwp} propose weight perturbation as a fairness-aware
generalization of SAM~\cite{foret2021sam}.
The original DP gap across groups is:
\begin{equation}
\begin{aligned}
  \Delta_{\text{DP}}(\bw) &= \max_g \hat{\mu}_g(\bw) - \min_g \hat{\mu}_g(\bw), \\
  \hat{\mu}_g &= \E_{(\bx,y)\in G_g}[f_\bw(\bx)].
\end{aligned}
\end{equation}
We identify that $\Delta_{\text{DP}}$ is theoretically susceptible to a
``fairness through uniformity'' failure mode: its optimum is reachable
by equalizing positive-prediction rates rather than correct predictions,
because the label $y$ is not conditioned upon.
Empirically this manifests unpredictably across architectures rather
than universally (\S\ref{sec:dp_diagnostic})---which is itself the
practical problem: a criterion whose failure mode cannot be anticipated
per-architecture is not a safe default.
We therefore adopt \textbf{FPR-disparity} as the default ascent/descent
criterion, conditioning only on real images ($y{=}0$), which removes the
theoretical precondition for the failure mode directly and is reliable
across every architecture we test:
\begin{equation}
\begin{aligned}
  \Delta_{\text{FPR}}(\bw) &= \max_g \hat{\mu}_g^{(0)}(\bw) - \min_g \hat{\mu}_g^{(0)}(\bw), \\
  \hat{\mu}_g^{(0)} &= \E_{\bx:\,y=0,\,g_i=g}[f_\bw(\bx)].
\end{aligned}
\label{eq:fpr_disp}
\end{equation}
\textbf{Step 1 (ascent)} perturbs toward worst-case disparity:
\begin{equation}
  \hat{\bw} = \bw + \rho \cdot
    \frac{\nabla_\bw \Delta_{\text{FPR}}(\bw)}{\|\nabla_\bw \Delta_{\text{FPR}}(\bw)\|}.
\end{equation}
\textbf{Step 2 (descent)} minimizes at $\hat{\bw}$ with a warmed-up penalty
$\lambda_t = \lambda\min(1, t/W)$:
\begin{equation}
  \bw \leftarrow \bw - \eta \nabla_\bw \bigl[
      \Lcls^{\text{gw}}(\hat{\bw}) + \lambda_t \Lfpr(\hat{\bw})
  \bigr],
\end{equation}
where $\Lcls^{\text{gw}}$ is BCE with label smoothing ($\varepsilon{=}0.05$)
and inverse-$\!\sqrt{}$ group-frequency weighting (upweighting demographically
rare samples within each batch).
Algorithm~\ref{alg:mwp} shows the full training loop.
\emph{Note:} the primary results in Tables~\ref{tab:main}--\ref{tab:utility}
use $\Delta_{\text{FPR}}$ with group-weighted BCE (Eq.~\ref{eq:fpr_disp}),
the default criterion in the released code.
Table~\ref{tab:dp_diagnostic} additionally reports the $\Delta_{\text{DP}}$
diagnostic (plain BCE, original weight-perturbation formulation) that motivated adopting
$\Delta_{\text{FPR}}$ in the first place (\S\ref{sec:dp_diagnostic}).
All four AI-Face detectors share the interface
$f_\theta(\texttt{data},\,\texttt{inference=True})\!\to\!\{\texttt{cls}\}$,
enabling a single unified loop.

\begin{algorithm}[t]
\caption{Weight-Perturbation Fine-Tuning ($\Delta_{\text{FPR}}$, the default criterion used for the primary results)}
\label{alg:mwp}
\begin{algorithmic}[1]
\REQUIRE Pretrained $f_{\bw_0}$, loader $\mathcal{D}$, $\rho$, $\lambda$, $\eta$, $T$, warmup $W$
\STATE Init optimizer (Adam, lr\,$\eta$, wd\,$10^{-5}$)
\STATE Init CosineAnnealingLR ($T_{\max}{=}T$, $\eta_{\min}{=}\eta/100$)
\FOR{$t = 1, \ldots, T$}
  \STATE $\lambda_t \leftarrow \lambda \cdot \min(1,\,t/W)$ \COMMENT{penalty warmup}
  \FOR{$(X, y, g) \sim \mathcal{D}_t$}
    \STATE \textbf{if} $|\mathrm{unique}(g)|{<}2$: \textbf{skip} batch \COMMENT{$\mathcal{D}_t$ is group-balanced, \S\ref{sec:impl}}
    \STATE $p \leftarrow \sigma(f_\bw(X).\texttt{cls})$
    \STATE $\Delta_{\text{FPR}} \leftarrow \max_g \hat{\mu}_g^{(0)} - \min_g \hat{\mu}_g^{(0)}$
    \STATE Backprop $\Delta_{\text{FPR}}$;\quad \textbf{first\_step}($\rho$)
    \STATE $p' \leftarrow \sigma(f_{\hat\bw}(X).\texttt{cls})$
    \STATE $\mathcal{L} \leftarrow \text{BCE}_{\text{gw}}(f_{\hat\bw}, y, g) + \lambda_t\,\Delta_{\text{FPR}}(p')$
    \STATE Backprop $\mathcal{L}$;\quad clip-grad\,(${\leq}1$);\quad \textbf{second\_step}()
  \ENDFOR
  \STATE scheduler.step()
\ENDFOR
\end{algorithmic}
\end{algorithm}

\subsection{Stage 2: Per-Group Youden-J Calibration}
\label{sec:cal}

After FPR-disparity weight perturbation, the checkpoint's AUC and intersectional fairness
have already improved for most architectures (Section~\ref{sec:results}),
but the decision boundary can still be sharpened per group.
We apply per-group threshold calibration~\cite{hardt2016equality} as a
zero-cost correction: for each group $g$, find the threshold $\tau_g^*$
that maximizes the Youden index,
\begin{equation}
  \tau_g^* = \arg\max_\tau \bigl[\text{TPR}_g(\tau) - \text{FPR}_g(\tau)\bigr],
\end{equation}
estimated on a held-out calibration split (50K training images, seed 99,
disjoint from weight-perturbation fine-tuning).
The threshold is applied as a logit shift: group $g$'s logit is shifted
by $\delta_g = -\log(\tau_g^* / (1-\tau_g^*))$, so the default 0.5 boundary
aligns with $\tau_g^*$ after shift.
No gradient updates are performed.

We evaluate three calibration modes:
\textbf{Youden-J} (per-group independent, recommended);
\textbf{eq-FPR} (shared optimal global FPR threshold); and
\textbf{eq-TPR} (shared optimal global TPR threshold)---the latter is
degenerate in isolation, since maximizing TPR alone with no FPR penalty
trivially drives the threshold toward predicting every image fake.
Youden-J is applied to the \emph{Stage-1-improved} checkpoint rather than to
the frozen pretrained baseline, which is the key distinction from post-hoc
calibration of unmodified detectors.
We call the combined two-stage pipeline \textbf{FYC} (Stage~1 + Stage~2).

\subsection{Model Suite and Implementation}
\label{sec:impl}

We apply weight perturbation to four AI-Face benchmark detectors:
\textbf{Xception}~\cite{chollet2017xception} and
\textbf{EfficientNet-B4}~\cite{tan2019efficientnet} (ERM baselines),
\textbf{DAG-FDD}~\cite{ju2023improving} (CVaR-based distributional training),
and \textbf{PG-FDD}~\cite{lin2024preserving} (explicit fairness head with
nested CVaR objectives).
All share the unified inference interface enabling a single weight-perturbation loop.

\mypar{Hyperparameters}
$T{=}15$ epochs, $\eta{=}10^{-5}$ (Adam, weight decay $10^{-5}$),
$\rho{=}0.05$, $\lambda{=}1.0$.
CosineAnnealingLR ($T_{\max}{=}T$, $\eta_{\min}{=}\eta/100$).
Penalty warmup $W{=}3$ epochs; label smoothing $\varepsilon{=}0.05$;
gradient clipping max-norm $1.0$.
Input: Resize $256{\times}256$, mean/std\,$=0.5$ (matching evaluation).
Per-epoch samples: 270K (Xception, DAG-FDD), 250K (EfficientNet),
130K (PG-FDD); drawn via inverse-$\!\sqrt{}$ group-frequency weighting
to ensure all six intersectional groups appear consistently.
PG-FDD uses batch size 16 (three-encoder overhead); others batch size 64.
Fine-tuning, measured per epoch with a regular (untuned)
\texttt{num\_workers}: 39\,min (Xception, directly measured), 39\,min
(DAG-FDD), 57\,min (EfficientNet), 46\,min (PG-FDD)---the latter three
scaled from their tuned times (22/22/32/26\,min with 48 dataloader
workers on a 128-core host) by Xception's measured
tuned-to-regular ratio ($1.77\times$)---${\approx}10$--15\,h total per
architecture over the full $T{=}15$ epochs on a single NVIDIA A100 GPU
(bf16 mixed precision, two forward/backward passes per step).

\section{Experiments}
\label{sec:experiments}

\subsection{Setup}
\label{sec:setup}

\mypar{Dataset}
AI-Face v2~\cite{aiface}: 1.3M training / 329,309 test images.
Six intersectional groups (Gender $\times$ Skin Tone: Light/Medium/Dark).
All evaluations use the full test split.
Baseline rows reproduce AI-Face's own published per-detector
results~\cite{aiface}; our raw weight-perturbation and calibrated
checkpoints are evaluated with our own \texttt{acc\_fairness()} pipeline
(global Youden-optimal threshold by default, \S\ref{sec:cal}), applied
identically across both of those stages.

\mypar{Metrics}
We report intersectional fairness using AI-Face's own metric
definitions~\cite{aiface}:
\textbf{FMEO} ($=\max(\Delta\text{TPR},\Delta\text{FPR})$);
\textbf{EO} ($=\Delta\text{TPR}+\Delta\text{FPR}$ $=$ FEO);
\textbf{FOAE} (max$-$min group accuracy);
\textbf{FPR} (global false-positive rate).
Per demographic axis (gender, skin tone, age) we report per-axis
\textbf{FMEO} and \textbf{FDP} (max$-$min positive prediction rate).
Utility: \textbf{AUC}, \textbf{Acc}, \textbf{AP}, \textbf{EER}.
Lower is better for FMEO, EO, FOAE, FPR, FDP, EER; higher for AUC, Acc, AP.

\mypar{Leveling-down}
We assess leveling down~\cite{mittelstadt2023unfairness} by comparing
overall Acc of the calibrated model to the pretrained baseline:
a method that reduces FMEO only by globally reducing accuracy
would show Acc$_{\text{method}}$ $<$ Acc$_{\text{base}}$.

\subsection{Comparison to the AI-Face Benchmark}
\label{sec:sota}

We first compare FYC against the full landscape of detectors benchmarked
on AI-Face v2, not just the two fairness-specialist architectures
(DAG-FDD, PG-FDD) detailed in \S\ref{sec:results}.
The AI-Face benchmark paper~\cite{aiface} itself reports results for 12
detectors on the same test set and \texttt{acc\_fairness()} protocol used
throughout this paper: three naive backbones (Xception, EfficientNet-B4,
ViT-B/16), three frequency-based methods (F3Net, SPSL, SRM), three
spatial-based methods (UCF, UnivFD, CORE), and three fairness-enhanced
methods (DAW-FDD, DAG-FDD, PG-FDD).
Table~\ref{tab:sota} reproduces their \emph{complete} per-detector,
per-axis breakdown (their Table 4 in full) alongside our two best FYC
results.
Our recipe---applied to a plain ERM backbone (Xception) or a CVaR-based
one (DAG-FDD)---wins 18 of the 21 reported rows outright against all 12
AI-Face-benchmarked detectors, including UnivFD and PG-FDD, the two
strongest baselines by AI-Face's own analysis.
This holds despite requiring only a pretrained checkpoint and
${\approx}10$--$15$\,h of post-hoc fine-tuning (39\,min per epoch for
Xception), versus full retraining for
every detector in Table~\ref{tab:sota}; the AI-Face authors report that
PG-FDD alone takes 7\,h\,20\,min \emph{per epoch} to
train~\cite{aiface}.
The three rows we do not win are, without exception, FDP (Demographic
Parity gap) on the Skin Tone, Age, and Intersectional axes---we win
Gender-FDP.
This is not a regression: FDP (demographic parity) is the metric the
objective we deliberately avoid (\S\ref{sec:mwp}) would target, so our
recipe does not optimize for it by design, and instead wins every
FEO (equalized odds) row---the criterion \S\ref{sec:discussion} argues is
the appropriate one for a detection task where the positive class is
the target, not a protected outcome.

\begin{table*}[t]
\centering
\caption{Full comparison to all 12 detectors benchmarked on
AI-Face v2~\cite{aiface} (their Table 4, reproduced in full, with
per-detector citations) plus our two best FYC results, each a single
run (\S\ref{sec:sota}).
FMEO/FDP/FOAE/FEO per demographic axis and Intersectional; Utility rows
below, including per-epoch training time from~\cite{aiface}.
\textbf{Bold} = best value per row across all 14 detectors.
FYC wins 18 of 21 fairness/utility rows (Time/Epoch excluded from
this tally); all three exceptions are FDP (\S\ref{sec:sota}).
X+FYC = Xception+FYC; D+FYC = DAG-FDD+FYC.}
\label{tab:sota}
\resizebox{\linewidth}{!}{%
{\renewcommand{\arraystretch}{0.72}\setlength{\tabcolsep}{3pt}%
\begin{tabular}{lccccccccccccccc}
\toprule
& & \textbf{Xcep}~\cite{chollet2017xception} & \textbf{EffB4}~\cite{tan2019efficientnet} & \textbf{ViT}~\cite{dosovitskiy2021vit} & \textbf{F3N}~\cite{qian2020f3net} & \textbf{SPSL}~\cite{liu2021spsl} & \textbf{SRM}~\cite{luo2021srm} & \textbf{UCF}~\cite{yan2023ucf} & \textbf{UnivFD}~\cite{ojha2023univfd} & \textbf{CORE}~\cite{ni2022core} & \textbf{DAW}~\cite{ju2023improving} & \textbf{DAG}~\cite{ju2023improving} & \textbf{PG}~\cite{lin2024preserving} & \textbf{X+FYC} & \textbf{D+FYC} \\
\midrule
\multirow{4}{*}{\rotatebox{90}{\textit{Skin}}}
 & FMEO$\downarrow$ & 8.84 & 8.30 & 6.26 & 19.94 & 8.06 & 10.00 & 17.33 & 2.58 & 10.78 & 14.12 & 6.55 & 6.47 & 3.34 & \textbf{1.24} \\
 & FDP$\downarrow$  & 9.75 & \textbf{6.18} & 7.73 & 12.88 & 9.38 & 10.90 & 12.58 & 8.56 & 10.32 & 10.71 & 8.62 & 9.75 & 9.48 & 8.14 \\
 & FOAE$\downarrow$ & 1.27 & 4.38 & 2.17 & 2.82 & 1.14 & 0.92 & 1.88 & 2.75 & 1.33 & 1.67 & 1.39 & 0.88 & \textbf{0.73} & 1.19 \\
 & FEO$\downarrow$  & 12.13 & 11.06 & 8.81 & 23.71 & 9.79 & 14.24 & 21.92 & 5.54 & 13.07 & 16.60 & 7.38 & 9.12 & 4.48 & \textbf{2.24} \\
\midrule
\multirow{4}{*}{\rotatebox{90}{\textit{Gender}}}
 & FMEO$\downarrow$ & 3.98 & 5.39 & 5.10 & 4.72 & 4.41 & 6.27 & 5.07 & 4.50 & 5.80 & 5.51 & 5.91 & 3.19 & 0.45 & \textbf{0.35} \\
 & FDP$\downarrow$  & 1.69 & 1.73 & 1.34 & 1.86 & 1.83 & 1.96 & 1.74 & 1.19 & 2.15 & 2.02 & 2.15 & 1.25 & 0.96 & \textbf{0.80} \\
 & FOAE$\downarrow$ & 0.98 & 1.49 & 1.80 & 1.13 & 1.04 & 1.77 & 1.45 & 1.62 & 1.39 & 1.33 & 1.42 & 1.07 & \textbf{0.21} & 0.33 \\
 & FEO$\downarrow$  & 4.14 & 5.86 & 6.03 & 4.87 & 4.53 & 6.78 & 5.51 & 5.41 & 5.93 & 5.70 & 6.07 & 3.70 & \textbf{0.59} & 0.68 \\
\midrule
\multirow{4}{*}{\rotatebox{90}{\textit{Age}}}
 & FMEO$\downarrow$ & 27.88 & 6.80 & 14.94 & 38.80 & 27.61 & 24.84 & 47.50 & 5.44 & 33.88 & 45.47 & 15.23 & 14.80 & 1.41 & \textbf{1.19} \\
 & FDP$\downarrow$  & 10.91 & 11.85 & 11.84 & 14.91 & 11.23 & 11.57 & 17.05 & 15.25 & 12.56 & 14.11 & \textbf{9.63} & 10.47 & 14.31 & 14.71 \\
 & FOAE$\downarrow$ & 7.27 & 2.86 & 6.84 & 10.12 & 7.27 & 6.52 & 11.65 & 3.79 & 8.76 & 11.88 & 5.53 & 5.01 & 1.17 & \textbf{1.09} \\
 & FEO$\downarrow$  & 42.22 & 10.30 & 30.80 & 55.03 & 40.94 & 38.53 & 67.55 & 14.15 & 48.73 & 64.38 & 30.18 & 29.59 & 2.45 & \textbf{2.24} \\
\midrule
\multirow{4}{*}{\rotatebox{90}{\textit{Inter.}}}
 & FMEO$\downarrow$ & 10.51 & 17.59 & 9.38 & 21.37 & 10.38 & 15.14 & 20.13 & 6.12 & 15.34 & 16.57 & 12.18 & 9.58 & 8.24 & \textbf{3.22} \\
 & FDP$\downarrow$  & 14.51 & \textbf{8.61} & 11.54 & 17.18 & 13.26 & 15.19 & 17.03 & 14.03 & 14.30 & 14.09 & 11.71 & 14.70 & 13.96 & 12.04 \\
 & FOAE$\downarrow$ & 2.54 & 8.46 & 4.93 & 4.87 & 2.46 & 4.00 & 3.54 & 6.29 & 2.78 & 3.55 & 4.04 & 3.06 & 1.84 & \textbf{1.83} \\
 & FEO$\downarrow$  & 24.32 & 25.11 & 27.44 & 47.78 & 21.68 & 30.11 & 43.38 & 20.26 & 28.84 & 33.12 & 26.30 & 18.35 & 10.36 & \textbf{6.11} \\
\midrule
\multirow{6}{*}{\rotatebox{90}{\textit{Utility}}}
 & AUC$\uparrow$ & 98.58 & 98.61 & 98.69 & 98.71 & 98.75 & 97.94 & 98.08 & 98.19 & 98.58 & 97.81 & 98.77 & 99.17 & 99.85 & \textbf{99.91} \\
 & ACC$\uparrow$ & 96.31 & 94.20 & 94.47 & 95.72 & 96.35 & 95.09 & 95.15 & 93.65 & 96.22 & 95.43 & 95.72 & 96.17 & 98.88 & \textbf{99.03} \\
 & AP$\uparrow$  & 99.35 & 99.54 & 99.57 & 99.45 & 99.36 & 99.17 & 99.27 & 99.40 & 99.36 & 99.02 & 99.50 & 99.69 & 99.93 & \textbf{99.96} \\
 & EER$\downarrow$ & 5.15 & 6.69 & 6.37 & 5.26 & 4.37 & 6.48 & 7.71 & 7.63 & 5.15 & 7.06 & 5.50 & 4.96 & 1.21 & \textbf{1.07} \\
 & FPR$\downarrow$ & 12.96 & 20.07 & 16.43 & 14.68 & 13.66 & 15.75 & 13.65 & 18.55 & 13.41 & 16.67 & 14.84 & 10.97 & 1.50 & \textbf{1.31} \\
 & Time/ep.$\downarrow$ & 1h15m & 2h25m & 2h40m & 1h18m & 1h20m & 3h10m & 5h05m & 4h00m & 1h16m & 1h25m & 1h17m & 7h20m & \textbf{39m} & \textbf{39m} \\
\bottomrule
\end{tabular}}}
\end{table*}

\subsection{Main Results}
\label{sec:results}

Having shown FYC's result against the full field (\S\ref{sec:sota}), we
now break down performance across all four of our evaluated architectures
individually.
Tables~\ref{tab:main} and~\ref{tab:utility} report baseline vs.\
FYC---the primary, released-code
configuration---for all four models.

\mypar{FPR-disparity weight perturbation improves fairness and utility together}
Unlike the DP objective, which degrades one of our four architectures
(\S\ref{sec:dp_diagnostic}), FPR-disparity weight perturbation improves
every architecture we test, with no exception.
With 3-seed means available for all four architectures, raw
fine-tuning alone already improves intersectional FMEO for
\emph{all four}, with no exception: Xception $-$74.2\%, DAG-FDD $-$75.9\%,
EfficientNet-B4 $-$34.6\%, PG-FDD $-$41.1\%.
AUC improves for all four at this same raw, pre-calibration
stage (Xception $+$1.29\,pp, DAG-FDD $+$1.15\,pp, EfficientNet-B4
$+$0.85\,pp, PG-FDD $+$0.57\,pp).
Unlike an earlier smaller-seed pass, EfficientNet-B4 shows no exception
at either metric once seed variance is properly averaged out ($\pm$0.01\,pp
AUC std across seeds)---the disparity criterion, not the architecture,
is what determines the raw-stage outcome (\S\ref{sec:discussion}).

\mypar{Youden-J adds further gains on top}
After per-group calibration, FMEO improves over baseline for
\emph{all four} architectures: Xception 10.51$\to$6.52$\pm$1.21
($-$38.0\%), EfficientNet-B4 17.59$\to$8.83$\pm$0.33 ($-$49.8\%),
DAG-FDD 12.18$\to$3.65$\pm$0.65 ($-$70.1\%), PG-FDD 9.58$\to$5.76$\pm$1.23
($-$39.8\%); all values 3-seed means.
Global FPR falls by 78.2--91.5\% across all four (Xception
12.96$\to$1.48$\pm$0.07; EfficientNet-B4 20.07$\to$3.39$\pm$0.14; DAG-FDD
14.84$\to$1.26$\pm$0.06; PG-FDD 10.97$\to$2.39$\pm$0.20).
AUC is maintained or improved for all four (Xception $+$1.29\,pp;
EfficientNet-B4 $+$0.85\,pp; DAG-FDD $+$1.15\,pp; PG-FDD $+$0.57\,pp).
EER improves for all four.

\mypar{FOAE is largely stable, unlike under DP}
FOAE (max$-$min group accuracy) does not worsen uniformly the way it does
under the DP diagnostic (\S\ref{sec:dp_diagnostic}): it improves for
Xception (2.54$\to$1.90$\pm$0.11), DAG-FDD (4.04$\to$1.78$\pm$0.05), and
EfficientNet-B4 (8.46$\to$6.57$\pm$0.13), and worsens only for PG-FDD
(3.06$\to$3.88$\pm$0.63).

\begin{table}[t]
\caption{Fairness metrics: Baseline vs.\ FYC
($+$FYC), $N=329{,}309$.
$+$FYC columns are mean$\pm$std over 3 seeds for all four architectures.
\bad{Red}: worse. \good{Teal}: better. \underline{Ul}: best baseline per row.
Per-axis FDP/FOAE/FEO breakdown for the two winning architectures is in
Table~\ref{tab:sota}.}
\label{tab:main}
\centering
\resizebox{\linewidth}{!}{%
{\renewcommand{\arraystretch}{0.72}\setlength{\tabcolsep}{2pt}%
\begin{tabular}{lcccccccc}
\toprule
\textbf{Metric}
  & \multicolumn{2}{c}{\textbf{Xception}}
  & \multicolumn{2}{c}{\textbf{EfficientNet-B4}}
  & \multicolumn{2}{c}{\textbf{DAG-FDD}}
  & \multicolumn{2}{c}{\textbf{PG-FDD}} \\
\cmidrule(lr){2-3}\cmidrule(lr){4-5}\cmidrule(lr){6-7}\cmidrule(lr){8-9}
& Base & $+$FYC & Base & $+$FYC & Base & $+$FYC & Base & $+$FYC \\
\midrule
\multicolumn{9}{l}{\textit{Intersectional ($\downarrow$)}} \\
FMEO & 10.51 & \good{6.52$\pm$1.21}  & 17.59 & \good{8.83$\pm$0.33}  & 12.18 & \good{3.65$\pm$0.65}  & \underline{9.58}  & \good{5.76$\pm$1.23} \\
EO          & 24.32 & \good{8.89$\pm$1.04}  & 25.11 & \good{11.03$\pm$0.48} & 26.30 & \good{6.17$\pm$0.61}  & \underline{18.35} & \good{10.39$\pm$2.93} \\
FOAE        & \underline{2.54} & \good{1.90$\pm$0.11} & 8.46  & \good{6.57$\pm$0.13}  & 4.04  & \good{1.78$\pm$0.05}  & 3.06              & \bad{3.88$\pm$0.63} \\
FPR         & 12.96 & \good{1.48$\pm$0.07}  & 20.07 & \good{3.39$\pm$0.14}  & 14.84 & \good{1.26$\pm$0.06}  & \underline{10.97} & \good{2.39$\pm$0.20} \\
\midrule
\multicolumn{9}{l}{\textit{Per-Axis FMEO ($\downarrow$)}} \\
Gen         & 3.98  & \good{0.52$\pm$0.08}  & 5.38  & \good{1.17$\pm$0.13}  & 5.91  & \good{0.42$\pm$0.06}  & \underline{3.19}  & \good{1.27$\pm$0.42} \\
Skin        & 8.84  & \good{2.44$\pm$0.71}  & 8.30  & \good{5.68$\pm$0.25}  & 6.55  & \good{1.33$\pm$0.37}  & \underline{6.46}  & \good{2.71$\pm$0.49} \\
Age         & 27.88 & \good{1.34$\pm$0.12}  & \underline{6.80} & \good{4.65$\pm$0.16}  & 15.23 & \good{1.27$\pm$0.22}  & 14.80 & \good{2.48$\pm$0.30} \\
\bottomrule
\end{tabular}}}
\end{table}

\begin{table}[t]
\caption{Utility metrics: Baseline vs.\ FYC.
$+$FYC rows are mean$\pm$std over 3 seeds for all four architectures.
\bad{Red}: degradation. \good{Teal}: improvement.
\underline{Underline}: best baseline per column.}
\label{tab:utility}
\resizebox{\linewidth}{!}{%
{\renewcommand{\arraystretch}{0.72}\setlength{\tabcolsep}{4pt}%
\begin{tabular}{lcccc}
\toprule
\textbf{Model / Setting} & \textbf{AUC$\uparrow$} & \textbf{Acc$\uparrow$} & \textbf{FPR$\downarrow$} & \textbf{EER$\downarrow$} \\
\midrule
\multicolumn{5}{l}{\textit{ERM baselines}} \\
Xception                  & 98.58 & \underline{96.31} & 12.96 & 5.15 \\
Xception + FYC   & \good{99.87$\pm$0.01} & \good{98.77$\pm$0.11} & \good{1.48$\pm$0.07}  & \good{1.27$\pm$0.08} \\
EfficientNet-B4           & 98.61 & 94.20 & 20.07 & 6.69 \\
EfficientNet + FYC & \good{99.46$\pm$0.01} & \good{96.44$\pm$0.11} & \good{3.39$\pm$0.14}  & \good{3.50$\pm$0.04} \\
\midrule
\multicolumn{5}{l}{\textit{Fairness-aware baselines}} \\
DAG-FDD                   & 98.77 & 95.72 & 14.84 & 5.50 \\
DAG-FDD + FYC    & \good{99.92$\pm$0.01} & \good{99.07$\pm$0.03} & \good{1.26$\pm$0.06}  & \good{1.03$\pm$0.04} \\
PG-FDD                    & \underline{99.17} & 96.17 & \underline{10.97} & \underline{4.96} \\
PG-FDD + FYC     & \good{99.74$\pm$0.04} & \good{97.67$\pm$0.25} & \good{2.39$\pm$0.20} & \good{2.29$\pm$0.20} \\
\bottomrule
\end{tabular}}}
\end{table}

\subsection{DP Diagnostic: Why FPR-Disparity?}
\label{sec:dp_diagnostic}

To motivate the criterion adopted in \S\ref{sec:mwp}, we ran the same
weight-perturbation loop with Jiang \etal's~\cite{jiang2023mwp} original Demographic Parity
(DP) disparity objective and plain BCE, keeping every other hyperparameter
fixed, and applied Stage 2 on top of the resulting checkpoints exactly as
for FPR-disparity.
Table~\ref{tab:dp_diagnostic} places both criteria side by side---baseline
$\to$ raw $\to$ calibrated---for direct comparison.

\mypar{DP is architecture-dependent, not uniformly catastrophic}
Raw DP-regularized weight perturbation \emph{improves} intersectional
FMEO for two of four architectures (Xception, DAG-FDD), is roughly
flat for EfficientNet-B4, and \emph{worsens} it only for PG-FDD
($\times$1.52).
Global FPR and AUC both improve for \emph{all four} architectures under
DP---so the classic ``predict fake broadly'' failure mode (elevated
global FPR) is absent even for PG-FDD, whose degradation is real but
concentrated elsewhere: skin-tone $\Delta$TPR rises sharply
(6.46$\to$14.18) while gender-axis FDP stays essentially flat.
The DP objective's theoretical vulnerability (\S\ref{sec:mwp}: its optimum
is reachable by equalizing positive-prediction rates rather than correct
predictions, since $y$ is not conditioned upon) is mathematically real,
but which architecture falls into that basin, and on which axis, is not
predictable from the objective alone---PG-FDD's own nested-CVaR fairness
machinery appears to interact destructively with an \emph{additional}
disparity objective layered on via weight perturbation, a combination the
other three architectures do not exhibit.

\mypar{FPR-disparity's raw stage never fails this way}
DP-regularized weight perturbation is not unusable---three of four
architectures tested here are competitive or better than baseline even
raw---but its architecture-dependence makes its raw stage an unreliable
default: a practitioner cannot know in advance whether a new checkpoint
lands in PG-FDD's failure basin (9.58$\to$14.55 FMEO, worse than
baseline) without running the diagnostic itself.
FPR-disparity's raw stage improves intersectional FMEO and AUC
for all four architectures with no exception (Table~\ref{tab:dp_diagnostic}).
Conditioning the disparity objective on real images only removes the
theoretical DP failure mode's precondition, which plausibly explains why
FPR-disparity's raw stage never lands in a PG-FDD-style failure basin
here, though we cannot rule one out beyond the four architectures
tested.
Stage 2 calibration, applied on top of the DP-perturbed checkpoints
exactly as for FPR-disparity, partially recovers PG-FDD's raw
failure (14.55$\to$10.26 FMEO); this is a useful safety net, but a
practitioner using DP would not know to apply it without first
diagnosing the failure, which is precisely the reliability gap
FPR-disparity closes at the source.

\begin{table}[h]
\caption{DP diagnostic: baseline $\to$ raw (pre-calibration) weight
perturbation, DP~\cite{jiang2023mwp} vs.\ FPR-disparity (ours), same
single seed (seed 1) both rows for a like-for-like comparison.
This motivated adopting FPR-disparity as FYC's default criterion.
\bad{Red}: worse than baseline. \good{Teal}: better.}
\label{tab:dp_diagnostic}
\centering
{\footnotesize\renewcommand{\arraystretch}{0.85}\setlength{\tabcolsep}{4pt}%
\begin{tabular}{@{}llcccc@{}}
\toprule
\textbf{Model} & \textbf{Criterion} & \multicolumn{2}{c}{\textbf{Baseline}} & \multicolumn{2}{c}{\textbf{Raw}} \\
& & FMEO$\downarrow$ & AUC$\uparrow$ & FMEO$\downarrow$ & AUC$\uparrow$ \\
\midrule
\multirow{2}{*}{Xception} & DP        & \multirow{2}{*}{10.51} & \multirow{2}{*}{98.58} & \good{5.63}  & \good{99.89} \\
                          & FPR-disp.\ (ours) & & & \good{3.01} & \good{99.85} \\
\multirow{2}{*}{EffB4}    & DP        & \multirow{2}{*}{17.59} & \multirow{2}{*}{98.61} & \good{13.75} & \good{99.52} \\
                          & FPR-disp.\ (ours) & & & \good{11.88} & \good{99.47} \\
\multirow{2}{*}{DAG-FDD}  & DP        & \multirow{2}{*}{12.18} & \multirow{2}{*}{98.77} & \good{3.10}  & \good{99.87} \\
                          & FPR-disp.\ (ours) & & & \good{3.32} & \good{99.91} \\
\multirow{2}{*}{PG-FDD}   & DP        & \multirow{2}{*}{9.58}  & \multirow{2}{*}{99.17} & \bad{14.55}  & \good{99.86} \\
                          & FPR-disp.\ (ours) & & & \good{4.13} & \good{99.80} \\
\bottomrule
\end{tabular}}
\end{table}

\begin{figure}[t]
\centering
\includegraphics[width=\linewidth]{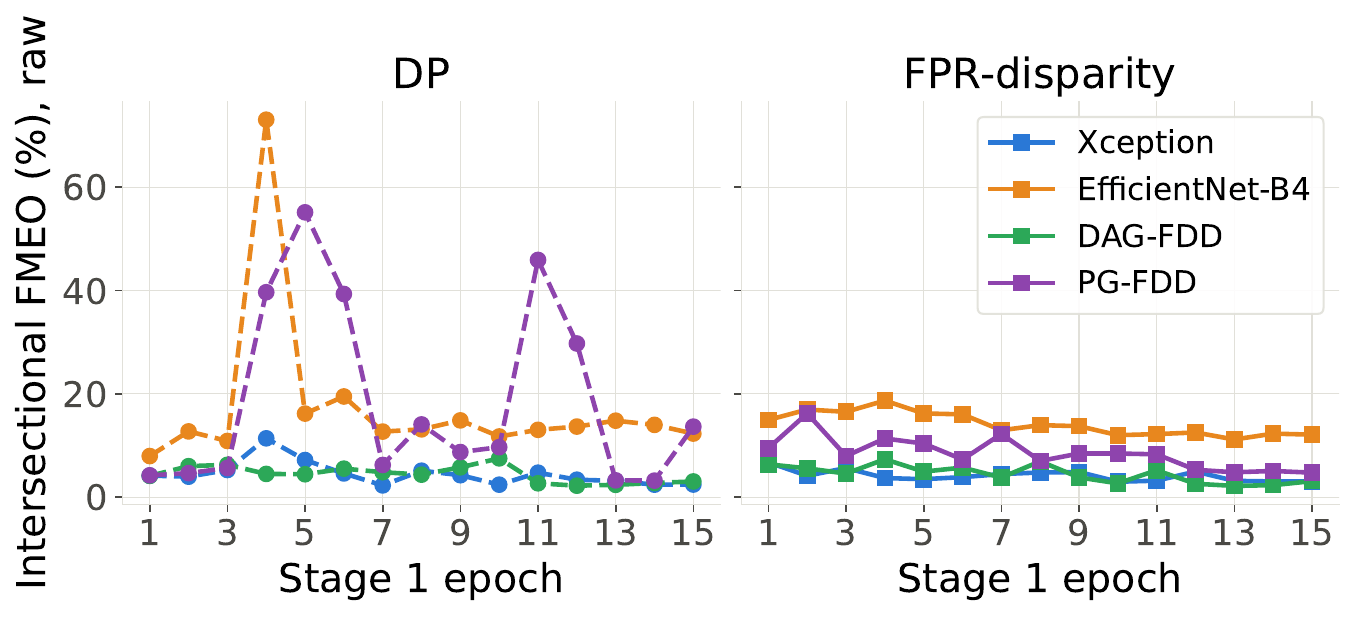}
\caption{Raw (pre-calibration) intersectional FMEO at every Stage 1
epoch, DP vs.\ FPR-disparity, for all four architectures (50k-image
random subsample of the test set, global-Youden threshold recomputed
per epoch).
DP is stable for Xception and DAG-FDD, spikes once for EfficientNet-B4,
and oscillates without converging for PG-FDD; FPR-disparity is stable
for all four.}
\label{fig:trajectory}
\end{figure}

\mypar{DP's instability holds across training, not just at one epoch}
Figure~\ref{fig:trajectory} tracks raw intersectional FMEO at every
Stage 1 epoch, on a 50k-image random subsample of the test set, for both
criteria on all four architectures.
FPR-disparity stays low and settles for every architecture.
Under DP, Xception and DAG-FDD stay low and stable throughout;
EfficientNet-B4 shows one sharp mid-training spike but is otherwise
contained; PG-FDD is the extreme case, swinging repeatedly between
roughly 3\% and 55\% with no sign of converging even by epoch 15,
confirming PG-FDD's raw-stage failure (Table~\ref{tab:dp_diagnostic}) is
not an artifact of a single checkpoint.

\mypar{Why Youden-J}
We use Youden-J throughout rather than a shared global equal-FPR or
equal-TPR threshold, because it is the only one of the three that
directly targets each group's own TPR$-$FPR tradeoff (\S\ref{sec:cal})
rather than forcing a single shared operating point across groups with
heterogeneous score distributions; we did not exhaustively ablate
against alternative calibration objectives such as
FRAPPÉ~\cite{tifrea2024frappe} or LinearPost~\cite{xian2024linearpost}.

\subsection{Calibration-Only Ablation}
\label{sec:cal_only}

To quantify weight perturbation's marginal contribution over calibration alone, we apply
Youden-J to the \emph{pretrained} checkpoint (no weight perturbation) and compare
to FYC in Table~\ref{tab:cal_only}.
This isolates whether the FPR and FMEO gains come from the
improved feature geometry (weight perturbation) or from threshold correction alone.

\begin{figure}[t]
\centering
\includegraphics[width=\linewidth]{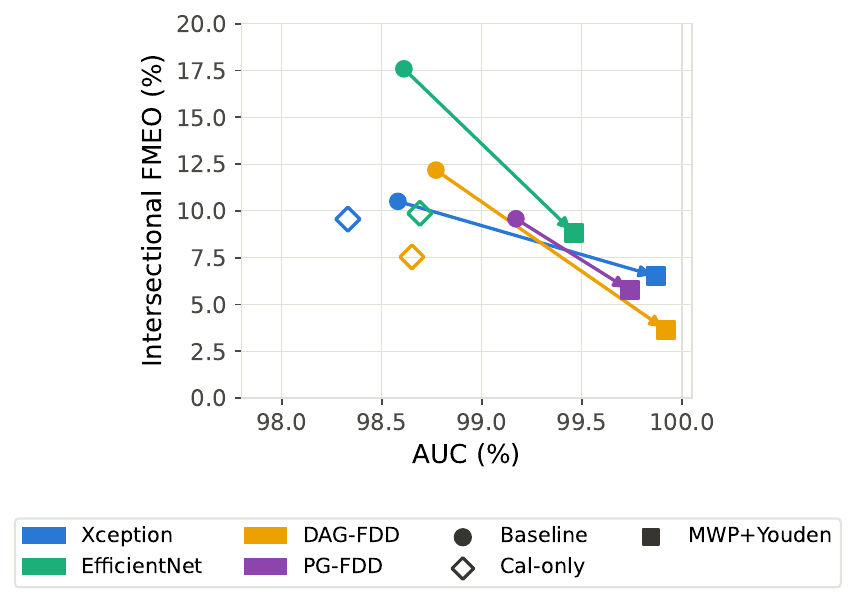}
\caption{Fairness--utility scatter (AUC vs.\ Intersectional FMEO).
Filled circles = baseline; filled squares = FPR-disparity
FYC; hollow diamonds = calibration-only (no weight perturbation). Arrows
connect baseline to FYC; ideal direction is lower-right
(higher AUC, lower FMEO).}
\label{fig:scatter}
\end{figure}

\begin{figure}[t]
\centering
\includegraphics[width=\linewidth]{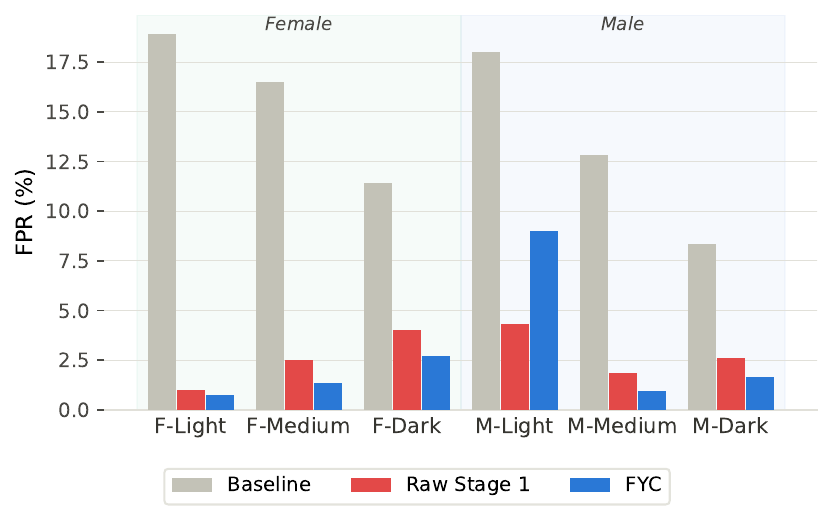}
\caption{Per-group FPR across the six intersectional groups
(Female/Male $\times$ Light/Medium/Dark skin tone) for Xception, train-seed 1.
Gray = pretrained baseline; red = raw Stage 1 (FPR-disparity, pre-calibration);
blue = FYC (Stage 1 + Stage 2). Calibration further reduces FPR for five of
six groups; M-Light is the one group where per-group Youden-J raises FPR
relative to the raw checkpoint in this particular seed, illustrating the
per-seed variance behind Table~\ref{tab:main}'s 3-seed means.}
\label{fig:per_group}
\end{figure}

With 3-seed means available for all four architectures, the result is
uniform: FYC wins \emph{every} metric for \emph{every} architecture
against calibration alone, with no exception (e.g.\ PG-FDD's Cal-only
\emph{worsens} FMEO relative to baseline, 9.58$\to$9.88, while
FYC reduces it to 5.76$\pm$1.23).
Weight perturbation is thus decisive for all four architectures we
test, not just some of them: calibration alone is never sufficient on
its own.
Figure~\ref{fig:scatter} visualizes this fairness--utility tradeoff
across all four architectures, and Figure~\ref{fig:per_group} breaks
Xception's result down to the six individual intersectional groups,
illustrating the per-seed variance behind Table~\ref{tab:main}'s 3-seed
means.

\begin{table}[h]
\caption{Calibration-only vs.\ FYC.
``Cal-only'' = Youden-J applied to the \emph{pretrained} checkpoint (no weight perturbation).
$+$FYC rows are mean$\pm$std over 3 seeds; Cal-only rows are single runs.
\textbf{Bold} = best value between Cal-only and FYC in each block.}
\label{tab:cal_only}
\resizebox{\linewidth}{!}{%
\begin{tabular}{lcccc}
\toprule
\textbf{Model / Setting} & \textbf{AUC$\uparrow$} & \textbf{Acc$\uparrow$} & \textbf{FMEO$\downarrow$} & \textbf{FPR$\downarrow$} \\
\midrule
\multicolumn{5}{l}{\textit{ERM baselines}} \\
Xception                    & 98.58 & 96.31 & 10.51 & 12.96 \\
Xception + Cal-only         & 98.33 & 96.12 &  9.56 &  7.39 \\
Xception + FYC     & \textbf{99.87$\pm$0.01} & \textbf{98.77$\pm$0.11} & \textbf{6.52$\pm$1.21} & \textbf{1.48$\pm$0.07} \\
EfficientNet-B4             & 98.61 & 94.20 & 17.59 & 20.07 \\
EfficientNet + Cal-only     & 98.69 & 93.69 &  9.87 &  6.76 \\
EfficientNet + FYC & \textbf{99.46$\pm$0.01} & \textbf{96.44$\pm$0.11} & \textbf{8.83$\pm$0.33} & \textbf{3.39$\pm$0.14}  \\
\midrule
\multicolumn{5}{l}{\textit{Fairness-aware baselines}} \\
DAG-FDD                     & 98.77 & 95.72 & 12.18 & 14.84 \\
DAG-FDD + Cal-only          & 98.65 & 95.39 &  7.54 &  7.10 \\
DAG-FDD + FYC      & \textbf{99.92$\pm$0.01} & \textbf{99.07$\pm$0.03} & \textbf{3.65$\pm$0.65} & \textbf{1.26$\pm$0.06} \\
PG-FDD                      & 99.17 & 96.17 &  9.58 & 10.97 \\
PG-FDD + Cal-only           & 97.72 & 94.73 &  9.88 &  9.68 \\
PG-FDD + FYC     & \textbf{99.74$\pm$0.04} & \textbf{97.67$\pm$0.25} & \textbf{5.76$\pm$1.23} & \textbf{2.39$\pm$0.20} \\
\bottomrule
\end{tabular}}
\end{table}

\section{Discussion}
\label{sec:discussion}

\mypar{Architecture matters for DP, not for FPR-disparity}
The DP penalty's theoretical vulnerability---its optimum is reachable by
equalizing positive-prediction rates rather than correct predictions,
since $y$ is not conditioned upon (\S\ref{sec:mwp})---does not manifest
as a uniform ``predict fake broadly'' failure mode (global FPR and AUC both
\emph{improve} under DP for all four); the damage is architecture- and
axis-specific instead, real but not predictable from family alone
(DAG-FDD, also CVaR-based, shows none of PG-FDD's vulnerability;
\S\ref{sec:dp_diagnostic}, \S\ref{sec:limitations}).
FPR-disparity removes the failure mode's precondition---conditioning on
real images only, approximating the FPR half of equalized
odds~\cite{hardt2016equality}---and architecture stops mattering: raw
fine-tuning improves AUC and FMEO/FPR for all four with no
exception, Youden-J adds further value (FPR $\downarrow$78.2--91.5\%,
FMEO $\downarrow$38.0--70.1\% vs.\ baseline), and beats
calibration alone for every architecture, including PG-FDD
(\S\ref{sec:cal_only}).
What architecture-dependence costs DP is the ability to trust the raw
stage \emph{before} calibrating; FPR-disparity's raw stage needs no such
trust, which is the practical argument for using it by default.

\mypar{FOAE is mixed, not uniformly worse}
Unlike DP---where FOAE worsens for EfficientNet-B4 and, severely, for
PG-FDD ($3.06\to11.93$ raw)---FYC \emph{improves} FOAE for three of four
architectures and worsens it only for PG-FDD (Table~\ref{tab:main}),
whose Acc still improves substantially alongside
(94.46$\to$97.67$\pm$0.25): minority accuracy is being lifted, not
majority accuracy degraded---leveling
down~\cite{mittelstadt2023unfairness} is not the concern, since
equalizing FPR and equalizing per-group accuracy are not always jointly
optimizable by the same threshold shift without group-balanced
data~\cite{mittelstadt2023unfairness}.

\mypar{Limitations and Future Directions}
\label{sec:limitations}
(1) We report DP-regularized weight perturbation only as a diagnostic
baseline (\S\ref{sec:dp_diagnostic}), not as a recipe we release, and we
still do not know \emph{why} PG-FDD in particular falls into its failure
basin; a released equalized-odds variant, wider $\rho$-sensitivity
sweeps beyond 0.05, and a targeted study of what makes PG-FDD vulnerable
are natural next steps.
(2) The DP-diagnostic magnitudes in Table~\ref{tab:dp_diagnostic} come
from a single run per model, so PG-FDD's specific number could shift
once seed variance is measured the way it was for our main results,
though our recommendation to prefer FPR-disparity does not hinge on that
figure; multi-seed replication would firm this up.
(3) The calibration-only ablation in Table~\ref{tab:cal_only} shows FYC
winning every metric we tested, but that has only been verified on our
four architectures, and extending it to a wider pool of detectors would
show how far the pattern generalizes.
(4) Group-specific Youden-J calibration requires labeled demographic
data at calibration time, which will not always be available;
noise-robust AUC-fairness~\cite{wu2025aucfairness} is a promising
label-free alternative worth exploring.

\section{Conclusion}
\label{sec:conclusion}

FYC, on a plain Xception backbone, beats all 12 AI-Face-v2
detectors~\cite{aiface} on 18 of 21 fairness/utility rows
(\S\ref{sec:sota}), at a fraction of the training
cost of any in-processing baseline. The disparity criterion, not the
architecture, determines whether the \emph{raw} stage is predictable: DP is
architecture-dependent and occasionally harmful
(\S\ref{sec:dp_diagnostic}), while FPR-disparity removes that
unpredictability and drives every gain in this paper, with Youden-J
calibration adding further zero-cost gains (\S\ref{sec:cal_only}).

{\small
\bibliographystyle{ieee_fullname}
\bibliography{egbib}

@String(CVPR  = {CVPR})

@String(ICCV  = {ICCV})

@String(ECCV  = {ECCV})

@String(NIPS  = {NeurIPS})

@String(ICLR  = {ICLR})

@String(ICML  = {ICML})

@inproceedings{aiface,
  author    = {L. Lin and Santosh and M. Wu and X. Wang and S. Hu},
  title     = {{AI-Face}: A Million-Scale Demographically Annotated {AI}-Generated Face Dataset and Fairness Benchmark},
  booktitle = CVPR,
  year      = {2025}
}

@inproceedings{jiang2023mwp,
  author    = {Z. Jiang and X. Han and H. Jin and G. Wang and R. Chen and N. Zou and X. Hu},
  title     = {Chasing Fairness Under Distribution Shift: A Model Weight Perturbation Approach},
  booktitle = NIPS,
  year      = {2023}
}

@inproceedings{foret2021sam,
  author    = {P. Foret and A. Kleiner and H. Mobahi and B. Neyshabur},
  title     = {Sharpness-Aware Minimization for Efficiently Improving Generalization},
  booktitle = ICLR,
  year      = {2021}
}

@inproceedings{chollet2017xception,
  author    = {F. Chollet},
  title     = {Xception: Deep Learning with Depthwise Separable Convolutions},
  booktitle = CVPR,
  year      = {2017}
}

@inproceedings{tan2019efficientnet,
  author    = {M. Tan and Q. V. Le},
  title     = {{EfficientNet}: Rethinking Model Scaling for Convolutional Neural Networks},
  booktitle = ICML,
  year      = {2019}
}

@inproceedings{trinh2021examining,
  author    = {L. Trinh and Y. Liu},
  title     = {An Examination of Fairness of {AI} Models for Deepfake Detection},
  booktitle = {IJCAI},
  year      = {2021}
}

@inproceedings{hardt2016equality,
  author    = {M. Hardt and E. Price and N. Srebro},
  title     = {Equality of Opportunity in Supervised Learning},
  booktitle = NIPS,
  year      = {2016}
}

@inproceedings{ju2023improving,
  author    = {Y. Ju and S. Hu and S. Jia and G. H. Chen and S. Lyu},
  title     = {Improving Fairness in Deepfake Detection},
  booktitle = {WACV},
  year      = {2024}
}

@inproceedings{lin2024preserving,
  author    = {L. Lin and X. He and Y. Ju and X. Wang and F. Ding and S. Hu},
  title     = {Preserving Fairness Generalization in Deepfake Detection},
  booktitle = CVPR,
  year      = {2024}
}

@inproceedings{szandala2025fairflip,
  author    = {T. Szandala and F. Ezzeddine and N. Rusin and S. Giordano and O. Ayoub},
  title     = {{Fair-FLIP}: Fair Deepfake Detection with Fairness-Oriented Final Layer Input Prioritising},
  booktitle = {arXiv preprint arXiv:2507.08912},
  year      = {2025}
}

@inproceedings{nadimpalli2022gbdf,
  author    = {A. V. Nadimpalli and A. Rattani},
  title     = {{GBDF}: Gender Balanced DeepFake Dataset Towards Fair DeepFake Detection},
  booktitle = {ICPR International Workshops and Challenges},
  year      = {2022}
}

@article{xu2024analyzing,
  author  = {Y. Xu and P. Terhorst and M. Pedersen and K. Raja},
  title   = {Analyzing Fairness in Deepfake Detection with Massively Annotated Databases},
  journal = {IEEE Transactions on Technology and Society},
  volume  = {5},
  number  = {1},
  pages   = {93--106},
  year    = {2024}
}

@inproceedings{mittelstadt2023unfairness,
  author    = {B. Mittelstadt and S. Wachter and C. Russell},
  title     = {The Unfairness of Fair Machine Learning: Levelling Down and Strict Egalitarianism by Default},
  booktitle = {arXiv preprint arXiv:2302.02404},
  year      = {2023}
}

@inproceedings{dosovitskiy2021vit,
  author    = {A. Dosovitskiy and L. Beyer and A. Kolesnikov and D. Weissenborn and X. Zhai and T. Unterthiner and M. Dehghani and M. Minderer and G. Heigold and S. Gelly and J. Uszkoreit and N. Houlsby},
  title     = {An Image is Worth 16{$\times$}16 Words: Transformers for Image Recognition at Scale},
  booktitle = ICLR,
  year      = {2021}
}

@inproceedings{qian2020f3net,
  author    = {Y. Qian and G. Yin and L. Sheng and Z. Chen and J. Shao},
  title     = {Thinking in Frequency: Face Forgery Detection by Mining Frequency-Aware Clues},
  booktitle = ECCV,
  year      = {2020}
}

@inproceedings{liu2021spsl,
  author    = {H. Liu and X. Li and W. Zhou and Y. Chen and Y. He and H. Xue and W. Zhang and N. Yu},
  title     = {Spatial-Phase Shallow Learning: Rethinking Face Forgery Detection in Frequency Domain},
  booktitle = CVPR,
  year      = {2021}
}

@inproceedings{luo2021srm,
  author    = {Y. Luo and Y. Zhang and J. Yan and W. Liu},
  title     = {Generalizing Face Forgery Detection with High-Frequency Features},
  booktitle = CVPR,
  year      = {2021}
}

@inproceedings{yan2023ucf,
  author    = {Z. Yan and Y. Zhang and Y. Fan and B. Wu},
  title     = {{UCF}: Uncovering Common Features for Generalizable Deepfake Detection},
  booktitle = ICCV,
  year      = {2023}
}

@inproceedings{ojha2023univfd,
  author    = {U. Ojha and Y. Li and Y. J. Lee},
  title     = {Towards Universal Fake Image Detectors that Generalize Across Generative Models},
  booktitle = CVPR,
  year      = {2023}
}

@inproceedings{ni2022core,
  author    = {Y. Ni and D. Meng and C. Yu and C. Quan and D. Ren and Y. Zhao},
  title     = {{CORE}: {CO}nsistent {RE}presentation Learning for Face Forgery Detection},
  booktitle = {CVPR Workshops},
  year      = {2022}
}

@inproceedings{cheng2025daid,
  author    = {H. Cheng and M.-H. Liu and Y. Guo and T. Wang and L. Nie and M. Kankanhalli},
  title     = {Fair Deepfake Detectors Can Generalize},
  booktitle = NIPS,
  year      = {2025}
}

@inproceedings{wu2025aucfairness,
  author    = {M. Wu and L. Lin and W. Zhang and X. Wang and Z. Yang and S. Hu},
  title     = {Preserving {AUC} Fairness in Learning with Noisy Protected Groups},
  booktitle = ICML,
  year      = {2025}
}

@article{brown2026fft,
  author    = {R. Brown and C. Russell},
  title     = {Toward Calibrated, Fair, and Accurate Deepfake Detection},
  journal   = {arXiv preprint arXiv:2606.09881},
  year      = {2026}
}

@inproceedings{ezeakunne2025icaart,
  author    = {U. Ezeakunne and C. Eze and X. Liu},
  title     = {Data-Driven Fairness Generalization for Deepfake Detection},
  booktitle = {Proceedings of the 17th International Conference on Agents and Artificial Intelligence (ICAART)},
  volume    = {3},
  pages     = {582--591},
  year      = {2025}
}

@article{hu2026aifacecompetition,
  author    = {S. Hu and L. Lin and others},
  title     = {The Competition of Fairness in {AI}-generated Face Detection: Methods and Results},
  journal   = {Machine Intelligence Research},
  volume    = {23},
  number    = {3},
  pages     = {501--527},
  year      = {2026}
}

@String(AAAI = {AAAI})

@article{xian2024linearpost,
  author    = {R. Xian and H. Zhao},
  title     = {A Unified Post-Processing Framework for Group Fairness in Classification},
  journal   = {arXiv preprint arXiv:2405.04025},
  year      = {2024}
}

@inproceedings{tifrea2024frappe,
  author    = {A. Tifrea and P. Lahoti and B. Packer and Y. Halpern and A. Beirami and F. Prost},
  title     = {{FRAPP\'E}: A Group Fairness Framework for Post-Processing Everything},
  booktitle = ICML,
  year      = {2024}
}

@inproceedings{liu2025racialbias,
  author    = {D. Liu and S. Wang and R. Peng and Y. Wang and S. Hu and W. Gao},
  title     = {Thinking Racial Bias in Fair Forgery Detection: Models, Datasets and Evaluations},
  booktitle = AAAI,
  year      = {2025}
}

@inproceedings{hou2025individualfairness,
  author    = {Y. Hou and F. Lin and H. Li and S. Hu},
  title     = {Rethinking Individual Fairness in Deepfake Detection},
  booktitle = {ACM Multimedia},
  year      = {2025}
}

@inproceedings{ding2025decoupling,
  author    = {F. Ding and W. Yi and Y. Zhou and X. He and H. Rao and S. Hu},
  title     = {Decoupling Bias, Aligning Distributions: Synergistic Fairness Optimization for Deepfake Detection},
  booktitle = {Proceedings of the IEEE/CVF Conference on Computer Vision and Pattern Recognition (CVPR)},
  year      = {2026}
}

@article{xue2024bmft,
  author    = {C. Xue and J. Yan and A. Dutt and F. Haider and X. Liu and S. McDonagh and S. A. Tsaftaris},
  title     = {{BMFT}: Achieving Fairness via Bias-based Weight Masking Fine-tuning},
  journal   = {arXiv preprint arXiv:2408.06890},
  year      = {2024}
}

@article{lu2026fairrel,
  author    = {X. Lu and J. Li and S. Zheng and H. Chen and Y. Guan},
  title     = {{FairReL}: Deepfake Detection using Fairness-Aware Representation Learning},
  journal   = {arXiv preprint arXiv:2608.28777},
  year      = {2026}
}

@article{zhang2026fairforensics,
  author    = {Y. Zhang and J. Wu and Z. Gao and L. Shen},
  title     = {{FairForensics}: Seeing Expressions and Parsing Demographics via Vision-Language Modeling for Generalizable Fair Deepfake Detection},
  journal   = {arXiv preprint arXiv:2608.01661},
  year      = {2026}
}
}

\end{document}